\RequirePackage{fix-cm}
\RequirePackage{fixltx2e}
\documentclass[twoside,english,aip, jcp,floatfix]{revtex4-1}
\usepackage{mathpazo}

\usepackage[T1]{fontenc}
\usepackage[latin9]{inputenc}
\usepackage[a4paper]{geometry}
\usepackage{xcolor}
\usepackage{babel}
\usepackage{amsmath}
\usepackage{amssymb}
\usepackage{graphicx}
\usepackage[unicode=true,pdfusetitle,
 bookmarks=false,
 breaklinks=true,pdfborder={0 0 0},pdfborderstyle={},backref=false,colorlinks=true]
 {hyperref}
\hypersetup{
 citecolor = blue, linkcolor = blue,  urlcolor  = blue}

\makeatletter

\providecommand{\tabularnewline}{\\}

\renewcommand{\textendash}{--}

\usepackage{orcidlink}

\makeatother

\begin{document}

\title{{\Large{}Can bin-wise scaling improve consistency and adaptivity
}\\
{\Large{}of prediction uncertainty for machine learning regression
?}}

\author{Pascal PERNOT \orcidlink{0000-0001-8586-6222}}

\affiliation{Institut de Chimie Physique, UMR8000 CNRS,~\\
Université Paris-Saclay, 91405 Orsay, France}
\email{pascal.pernot@cnrs.fr}

\selectlanguage{english}%
\begin{abstract}
\noindent Binwise Variance Scaling (BVS) has recently been proposed
as a \emph{post hoc} recalibration method for prediction uncertainties
of machine learning regression problems that is able of more efficient
corrections than uniform variance (or temperature) scaling. The original
version of BVS uses uncertainty-based binning, which is aimed to improve
calibration conditionally on uncertainty, i.e. \emph{consistency}.
I explore here several adaptations of BVS, in particular with alternative
loss functions and a binning scheme based on an input-feature ($X$)
in order to improve \emph{adaptivity}, i.e. calibration conditional
on \emph{X}. The performances of BVS and its proposed variants are
tested on a benchmark dataset for the prediction of atomization energies
and compared to the results of isotonic regression. 
\end{abstract}
\maketitle

\section{Introduction}

\noindent \emph{Post hoc} recalibration of machine learned (ML) prediction
uncertainty for regression problems is currently an essential step
to increase the reliability of uncertainty quantification (UQ)\citep{Tran2020}.
\emph{Post hoc} recalibration methods include, for instance, temperature
scaling\citep{Guo2017,Kuleshov2018,Levi2022}, isotonic regression\citep{Busk2022}
and conformal inference\citep{Angelopoulos2021,Hu2022}. These methods
focus either on \emph{average calibration} or on \emph{consistency}
(i.e. calibration \emph{conditional} on uncertainty)\citep{Pernot2023c_arXiv}.
Until now, \emph{adaptivity}, i.e. calibration conditional on input
features\citep{Angelopoulos2021}, has surprisingly been left out
of the frame of \emph{post hoc} ML-UQ, despite its essential role
in UQ reliability for the end user\citep{Reiher2022,Pernot2023c_arXiv}.
It is important to stress out that prediction uncertainties with a
perfect consistency are not necessarily reliable across the range
of input features\citep{Pernot2023_Arxiv}. Consistency and adaptivity
are two facets of calibration that should be simultaneously validated
in order to approach the reliability of individual predictions.

Recently, a \emph{post hoc} recalibration method based on \emph{binwise
variance scaling} (BVS) has been proposed by Frenkel and Goldberger\citep{Frenkel2023},
adapted from a similar method for classification problems\citep{Frenkel2021}.
In this approach, a set of prediction uncertainties of size $M$ is
sorted and split into $N$ equal-size bins. For each bin, a local
uncertainty scaling factor is estimated to ensure calibration. These
scaling factors are then applied to newly predicted uncertainties.
In this setup based on uncertainty binning, the BVS approach aims
to improve consistency. Indeed, BVS has been shown to perform better
than uniform variance scaling for medical images regression tasks
on the basis of consistency metrics (ENCE, UCE)\citep{Frenkel2023}. 

In this short study I try to check if BVS can be used to improve both
consistency and adaptivity beyond the trivial effect of average calibration.
The next section (Sect.\,\ref{sec:Methods-1}) presents the BVS method,
alternative loss functions and the scores or calibration metrics used
to evaluate their performances. Application to the QM9 dataset of
atomization energies is presented in Sect.\,\ref{sec:Results}, leading
to the discussion of the results and conclusions in Sect.\,\ref{sec:Discussion-and-conclusion}.

\section{Methods\label{sec:Methods-1}}

\noindent Frenkel and Goldberger\citep{Frenkel2023} describe two
approaches to BVS estimation. One is based on the equality of mean
squared errors with mean squared uncertainty, and the other is based
on the maximum likelihood solution, which is consistent with the\emph{
z-}scores\emph{ }based approach favored in my earlier studies.\citep{Pernot2022b,Pernot2023c_arXiv}
As both approach appear to produce similar results, I focus here on
the second one.

\subsection{\emph{Z}-scores based BVS}

\noindent Let us consider a set of $M$ prediction errors and uncertainties
$\left\{ E_{i},u_{i}\right\} _{i=1}^{M}$. The dataset is ordered
by increasing uncertainty values and split into $N_{B}$ equal-size
bins $B_{1},\ldots,B_{N_{B}}$. The \emph{local} calibration within
bin $i$ can be assessed by the $z$-scores mean square (ZMS)\textcolor{orange}{\citep{Pernot2023c_arXiv}},
that should be equal to 1, i.e.
\begin{equation}
v_{i}=<Z_{j}^{2}>_{j\in B_{i}}=1\label{eq:LZV-1-1}
\end{equation}
where $Z_{j}=E_{j}/u_{j}$. 

A corrective scaling factor is thoroughly obtained as 
\begin{equation}
s_{i}=v_{i}^{-1/2}\label{eq:LZVscale-2-1}
\end{equation}
to be applied to uncertainties falling within the limits of the corresponding
bin
\begin{equation}
u_{j\in B_{i}}\rightarrow u_{s,j\in B_{i}}=s_{i}u_{j\in B_{i}}
\end{equation}
where $u_{s}$ denotes the scaled uncertainty.

The use of $u$ as a binning variable is adapted to improve consistency,
but other variables can be used to target adaptivity, as shown below. 

\subsubsection*{How many bins ?}

\noindent BVS is typically based on an equal-size binning scheme,
but the optimal bin number has to be defined. Reduction of the bin
number to $N_{B}=1$ leads to the so-called \emph{temperature scaling}
method\citep{Guo2017}. Temperature scaling is able to ensure \emph{average
calibration} but not consistency.\citep{Pernot2017,Pernot2017b,Pernot2022a}
At the opposite, using one scaling factor per data point, $N_{B}=M$,
leads to the singular solution $u_{s,j}=|E_{j}|$, the so-called \emph{oracle}
used in confidence curves\citep{Pernot2022c}. This scheme is useless,
as having null-width intervals it cannot be applied to unseen uncertainties.
In their applications, Frenkel and Goldberger\citep{Frenkel2023}
used $N_{B}=15$ without further justification. One might want to
use more bins to account for possible small-scale consistency/adaptivity
defects, but care has to be taken about over-parameterization. As
BVS is a fast method, a systematic study over a wide range of bin
numbers can be designed to orient the choice of the best $N_{B}$
value (Sect.\,\ref{subsec:Optimal-BVS-binning}).

\subsection{Score functions\label{subsec:Score-functions}}

\noindent Score functions are used to estimate the effect of scaling
on the quality of the resulting uncertainties, but also to enable
the estimation of optimal scaling factors by minimization of adapted
loss functions. 

Scores based on local statistics are based on a binning scheme which
is independent of the BVS binning scheme. In the following, I use
a partition of the binning variables into $N_{S}$ equal-size bins,
$B_{S,1},\ldots,B_{S,N_{S}}$. Other partitions have been considered
(e.g. adaptive binning based on equal-with bins, see Appendix B in
Ref. \citep{Pernot2023_Arxiv}), but they do not bring any notable
improvement to the results presented below.

\subsubsection{Score functions based on mean squared z-scores}

\noindent Scores are defined to measure the closeness of the\emph{
z}-scores mean squares (ZMS) to 1. For average calibration, one defines
\begin{equation}
S_{cal}=|\ln<Z^{2}>|
\end{equation}
where the average is taken over the full dataset. The logarithm accounts
for the fact that we are considering a scale statistic (a ZMS of 2
entails a correction of the same amplitude as a ZMS of 0.5). For perfect
average calibration, one should have $S_{cal}=0$. 

Similarly, one can define a mean calibration error
\begin{equation}
S_{x}=\frac{1}{N_{S}}\sum_{i=1}^{N_{S}}|\ln<Z_{j}^{2}>_{j\in B_{S,i}}|
\end{equation}
where $i$ runs over the $N_{S}$ bins and $x$ is the variable used
to define those bins. $S_{x}$ has to be as small as possible but
cannot be expected to reach 0 (see Sect.\,\ref{subsec:Target-values-for}
for the estimation of a target value).

A combination of the $S_{x}$ scores can be used to design loss functions
in order to achieve specific goals. For instance, consistency can
be targeted by minimizing
\begin{equation}
S_{con}=S_{cal}+S_{u}
\end{equation}
while adaptivity can be targeted with
\begin{equation}
S_{ada}=S_{cal}+\sum S_{X_{i}}
\end{equation}
where $X_{i}$ is an input feature or an adequate proxy. The combination
of all these scores
\begin{equation}
S_{tot}=S_{cal}+S_{u}+\sum S_{X_{i}}
\end{equation}
measures consistency and adaptivity. Note that $S_{cal}$ is always
included, as it is a basic requirement for proper calibration.

\subsubsection{Negative Log Likelihood}

\noindent Another option to optimize the BVS factors is to minimize
the average negative log likelihood function (NLL)\citep{Frenkel2023}
\begin{align}
NLL & =\frac{1}{2M}\sum_{i=1}^{M}\left(\frac{E_{i}^{2}}{u_{i}^{2}}+\ln u_{i}^{2}+\ln2\pi\right)\\
 & =\frac{1}{2}\left(<Z^{2}>+<\ln u^{2}>+\ln2\pi\right)
\end{align}
which combines an \emph{average calibration} term based on the mean
squared \emph{z}-scores, and a \emph{sharpness} term driving the uncertainties
towards small values\citep{Gneiting2007a}, hence preventing the minimization
of $<Z^{2}>$ by arbitrary large uncertainties. Frenkel and Goldberger\citep{Frenkel2023}
have shown that the scaling factors defined in Eq.\,\ref{eq:LZVscale-2-1}
minimize the NLL.

\subsubsection{ENCE }

\noindent For reference, let us consider also a statistic used by
Frenkel and Goldberger\citep{Frenkel2023}, namely the Expected Normalized
Calibration Error
\begin{equation}
ENCE=\frac{1}{N_{S}}\sum_{i=1}^{N_{S}}\frac{|MV_{i}^{1/2}-MSE_{i}^{1/2}|}{MV_{i}^{1/2}}
\end{equation}

\noindent where $MV_{i}=\frac{1}{|B_{S,i}|}\sum_{j\in B_{S,i}}u_{j}^{2}$
and $MSE_{i}=\frac{1}{|B_{S,i}|}\sum_{j\in B_{S,i}}E_{j}^{2}$. It
expresses the calibration error as a percentage and should be as close
to 0 as possible. A specific target value is defined below. 

Similarly to the $S_{x}$ scores, one can define $ENCE_{u}$ as a
consistency metric and $ENCE_{X_{i}}$ as adaptivity metric(s) by
using adapted binning schemes. Note that the extension of mean calibration
error to alternative binning schemes is being used in the literature
to measure \emph{adversarial} calibration\citep{Chung2021}. 

\subsection{Target values for the NLL, $S_{x}$ and $ENCE_{x}$ scores\label{subsec:Target-values-for}}

\noindent For a given dataset, the best values reachable by statistics
such as the $S_{x}$ or ENCE are not 0, as they result from the mean
of a sample of positive values. For an illustration about ENCE, see
for instance Pernot\citep{Pernot2023a_arXiv}. The optimal value depends
on the size of the dataset and on $N_{S}$. In order to get an estimation,
one needs a dataset offering compatibility between errors and uncertainties.
As one cannot easily infer perfect uncertainties, one has to generate
pseudo-errors from the available uncertainties, using for instance
a zero-centered normal distribution: $\tilde{E_{i}}\sim N(0,u_{i})$
and calculate the scores using $\tilde{E}$ and $u$. Sampling a large
number (e.g. $10^{3}$) of such generated datasets provides an estimate
of the optimal score values and of their uncertainties.\citep{Pernot2022c,Rasmussen2023}
These values are denoted as simulated (``Simul'') in the following. 

It has to be noted that for the NLL the simulated value is not necessarily
smaller than the actual values because the data are modified. In this
case, one has thus to consider the amplitude of the difference between
the actual NLL and its simulated value.\citep{Rasmussen2023}

\subsection{Validation metrics\label{subsec:Validation-metrics}}

\noindent The scores presented above enable to compare between methods
but do not directly enable to validate calibration. Estimation of
confidence intervals (CI) on the local ZMS values provides the basis
for a validation metric\citep{Pernot2022a,Pernot2023c_arXiv}: for
a perfectly calibrated dataset, the fraction of binned statistics
with a confidence interval containing the target value should be close
to the coverage probability of the CIs. Namely, about 95\,\% of the
local ZMS values should have their 95\,\% CI containing the target
value. Let us denote this fraction of ``valid'' intervals by $f_{v}$.
In practice, one should not expect to recover exactly $f_{v}=0.95$,
and a CI for $f_{v}$ has also to be estimated from the binomial distribution
to account for the limited number of bins\citep{Pernot2022a}. The
value of $f_{v}$ depends on the binning variable $x\in\left\{ u,X_{i}\right\} $
and is indexed accordingly as $f_{v,x}$.

For a correct estimation of $f_{v,x}$, it is important to ensure
a good balance between the bin size (estimation of local ZMS values
and their CIs by bootstrapping, as described in Ref.\,\citep{Pernot2023c_arXiv})
and the number of bins (estimation of $f_{v,x}$ and its CI). Using
a bin size of $M^{1/2}$ seems to be a rational choice\citep{Watts2022},
which might limit the use of $f_{v}$ to datasets with at least $10^{4}$
points. The 95\,\% CIs on local ZMS values are estimated by the BCa
bootstrapping method\citep{DiCiccio1996} with 1500 draws, and the
number of bins is set to $N_{S}=100$. 

\section{Results\label{sec:Results}}

\noindent To illustrate the methods presented above, I use a set of
data produced and presented by J. Busk \emph{et al.}\citep{Busk2022}.
The QM9 dataset contains predicted atomization energies ($V$) and
uncertainties ($u_{V}$), reference values ($R$), and molecular formulas
as input feature. These data are transformed to\textcolor{violet}{{}
}$C=\left\{ X_{1},X_{2},E=R-V,u=u_{V}\right\} $, where $X_{1}$ is
the \emph{molecular mass} and $X_{2}$ the \emph{fraction of hetero-atoms}
generated from the formulas\citep{Pernot2023_Arxiv,Bakowies2021}.
$X_{1}$ and $X_{2}$ are used here as complementary proxies for the
molecular formula. 

In the original study, the uncertainties were recalibrated by isotonic
regression\citep{Busk2022} and were later shown to reach a good consistency
while having adaptivity problems\citep{Pernot2023c_arXiv}. It is
not pertinent to test BVS on these transformed data (mostly because
of the stratification induced by isotonic regression)\citep{Pernot2023b_arXiv},
and J. Busk kindly provided me with the validation set ($M=10000$)
and test set ($M=13885$) of \emph{uncalibrated} predictions used
for their isotonic regression study. It has to be noted that the $X_{1}$
and $X_{2}$ variables are stratified, i.e. they achieve a finite
number of values over which the data points are distributed. This
can have an impact when they are used to define bins\citep{Pernot2023b_arXiv}. 

This datasets enable the direct comparison of BVS and isotonic regression
performances. BVS factors are estimated on the validation set (denoted
below as \emph{training} set) and their transferability is assessed
on the test set. Optimization of scaling factors provided by Eq.\,\ref{eq:LZVscale-2-1}
was attempted for $u$- and $X_{1}$-based binning schemes and for
the NLL, $S_{tot}$, $S_{con}$ and $S_{ada}$ loss functions. In
all the cases, the scaling factors were initialized by Eq.\,\ref{eq:LZVscale-2-1}
and optimized by a global optimizer, followed by a local optimizer
until convergence. When the score of the optimized solution did not
improve on the initial values (as expected for the NLL loss), the
latter were kept as the optimum. 

\subsection{Unscaled data\label{subsec:Unscaled-data}}

\noindent Average calibration of the training set is assessed by comparing
the ZMS to 1. One gets $<Z^{2}>=0.31(1)$, revealing a mismatch between
errors and uncertainties. This corresponds to the calibration score
$S_{cal}=1.17$ in Table\,\ref{tab:Scores-and-validation-20}. Similar
values are obtained for the test set ($<Z^{2}>=0.32(1)$, $S_{cal}=1.13$).
These $S_{cal}$ values have to be compared to their simulated target,
0. The other scores for both uncalibrated sets are also comparable.
Globally, the datasets are not biased, with $<Z>\le0.02$.

\begin{figure}[t]
\noindent \begin{centering}
\includegraphics[width=0.45\textwidth]{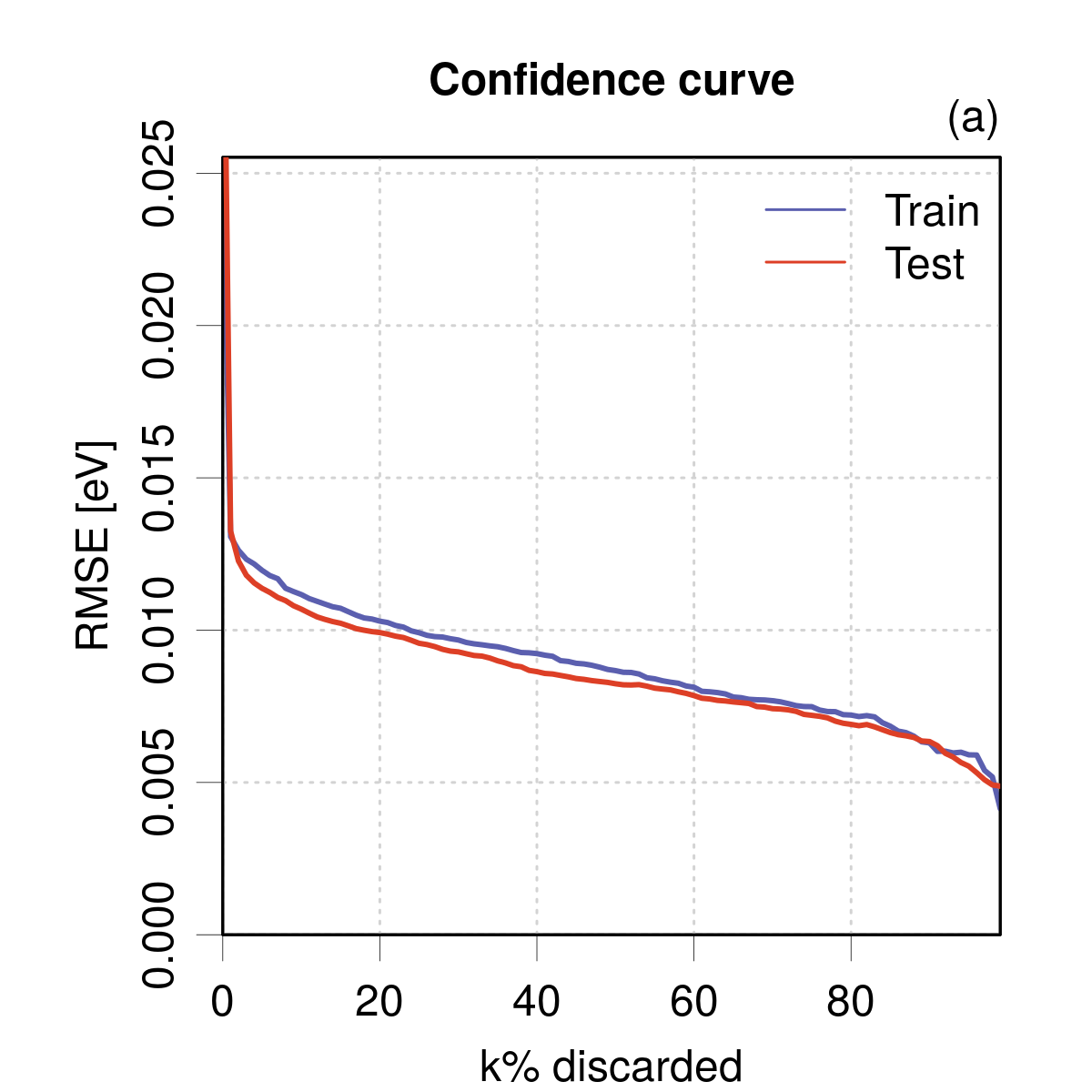}\includegraphics[width=0.45\textwidth]{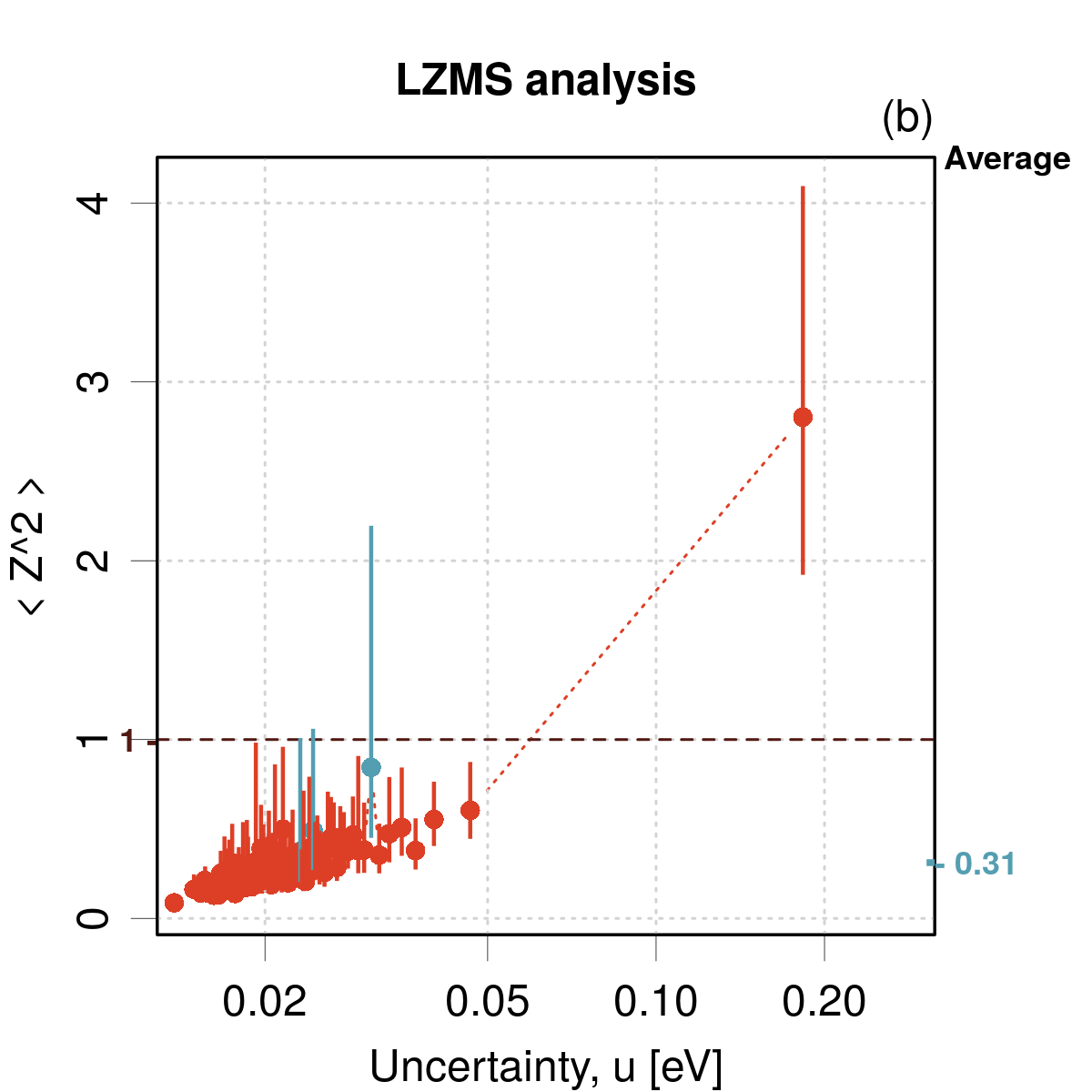}
\par\end{centering}
\noindent \begin{centering}
\includegraphics[width=0.45\textwidth]{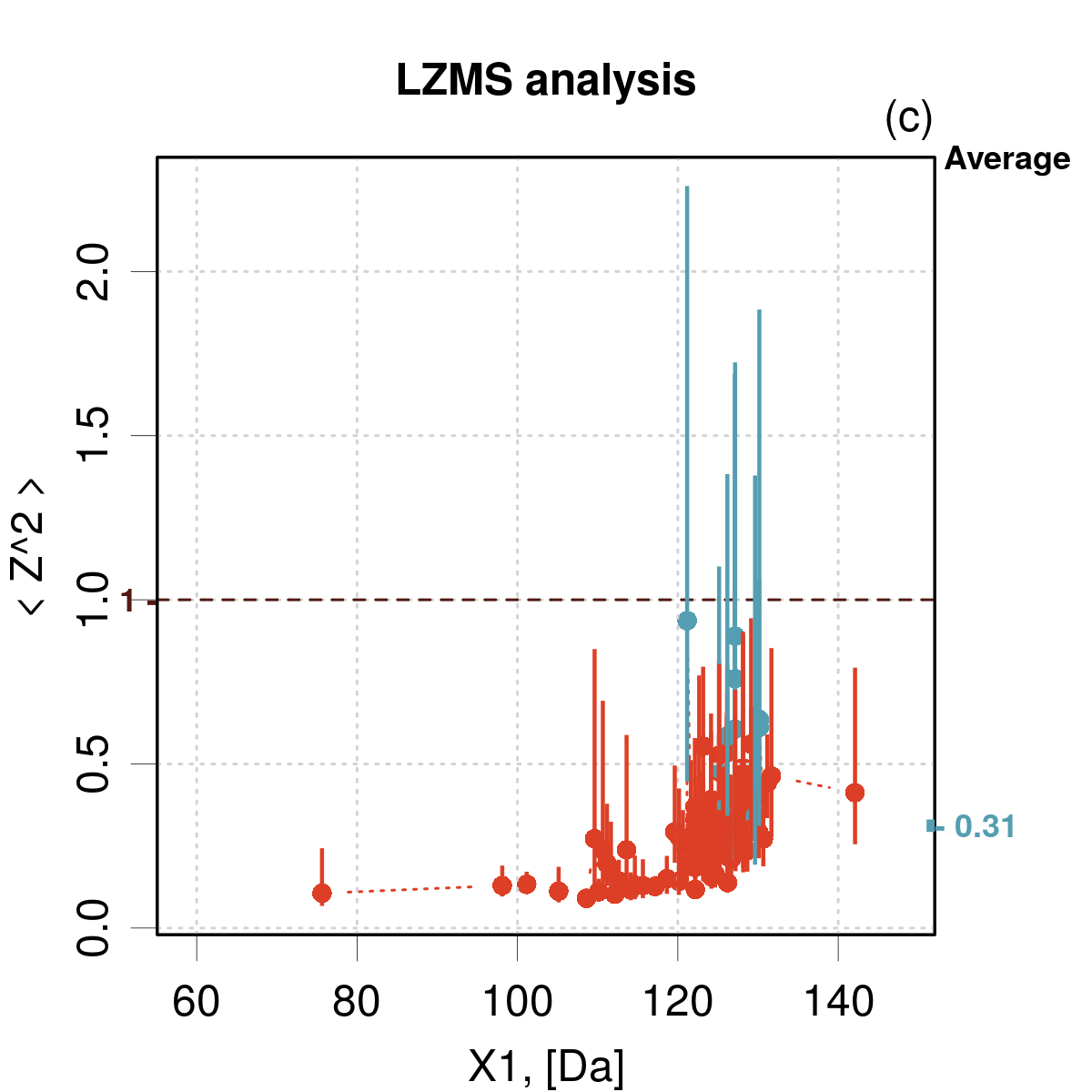}\includegraphics[width=0.45\textwidth]{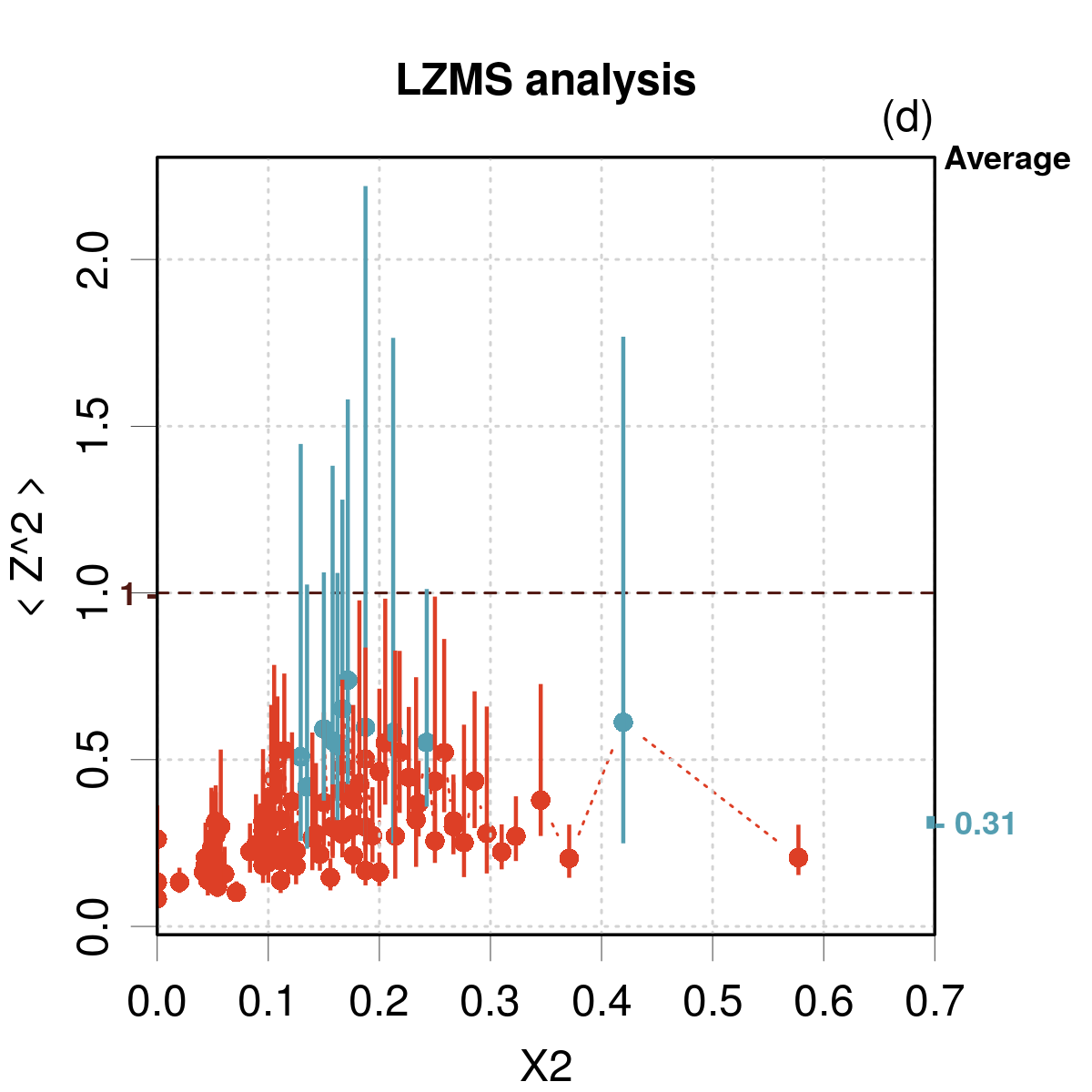}
\par\end{centering}
\caption{\label{fig:Confidence-curves}Analysis of the unscaled datasets: (a)
confidence curves for the training and test sets; (b-d) LZMS analysis
for the training set against $u$, $X_{1}$ and $X_{2}$ (the red
intervals do not contain the target value).}
\end{figure}

It is always possible to correct average calibration by a uniform
scaling of the uncertainties, but if one wants to reach consistency,
the uncertainties need to have some level of positive correlation
with the errors. This can be checked by plotting a \emph{confidence
curve}, displaying the RMS of error sets iteratively pruned from the
points with the largest uncertainties.\citep{Pernot2022b,Pernot2022c}
Note that confidence curves are insensitive to the scale of errors
and uncertainties. A continuously decreasing curve reveals a sane
situation where large errors are associated with large uncertainties,
and where one might hope to improve the situation with post-hoc calibration.
This is the case for the studied datasets {[}Fig.\,\ref{fig:Confidence-curves}(a){]},
for which the confidence curves show a sharp fall-off followed by
a continuous decrease. The sharp feature at the beginning shows that
the largest uncertainties are correctly associated to very large errors.

Consistency can be assessed on a local Z-mean-squares (LZMS) plot\citep{Pernot2023c_arXiv}
{[}Fig.\,\ref{fig:Confidence-curves}(b){]}. The positive slope of
the LZMS values as a function of $u$ shows that a uniform scaling
will not be sufficient to bring all the CIs to enclose the $<Z^{2}>=1$
line. The outlying point at high uncertainty corresponds to the sharp
feature observed in the confidence curve. The LZMS analysis shows
also that a monotonous scaling (linear or isotonic) can probably be
effective to restore consistency (as shown by Busk \emph{et al.} \citep{Busk2022}).
The LZMS plot for the test set is very similar (not shown). Analysis
of adaptivity by LZMS plots against $X_{1}$ and $X_{2}$ {[}Fig.\,\ref{fig:Confidence-curves}(c-d){]}
does not present the regularity observed for the consistency analysis.
The LZMS statistics are more locally dispersed, and it is clear that
a uniform or linear scaling of $u$ has little chance to be fully
efficient.

\subsection{Optimal BVS binning\label{subsec:Optimal-BVS-binning}}

\begin{figure}[t]
\noindent \begin{centering}
\includegraphics[width=0.9\textwidth]{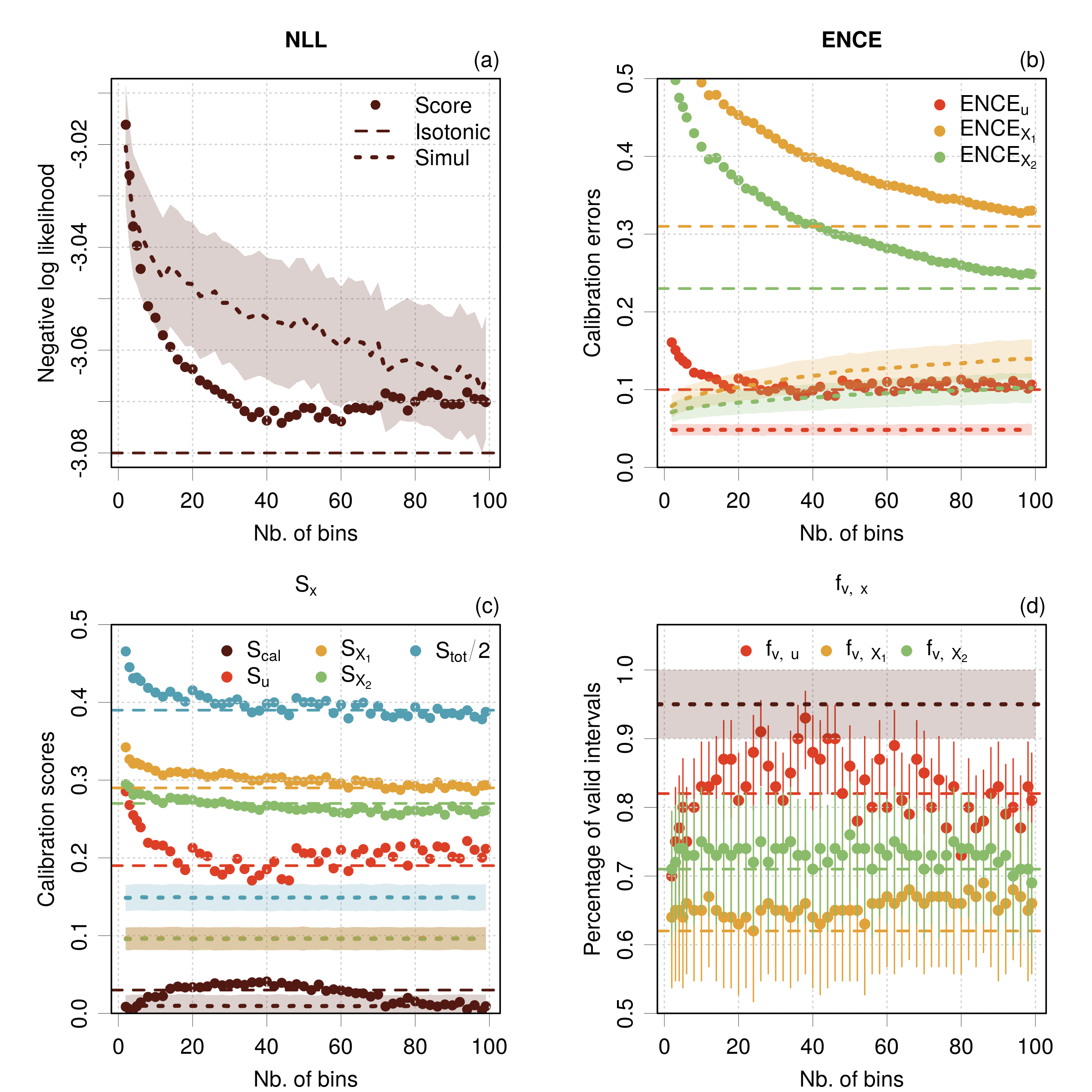}
\par\end{centering}
\caption{\label{fig:optBin}Dependence of the BVS scores on the test set with
the number of bins. The dashed horizontal lines depict the values
for isotonic regression and the dotted lines the values for simulated
datasets. The gray area in panel (d) corresponds to a 95\,\% CI on
the $f_{v,x}$ values for simulated data. $S_{tot}$ and its isotonic
and simulated references are scaled by a factor 0.5 for legibility.}
\end{figure}

\noindent The performance of the BVS scores resulting from Eq.\,\ref{eq:LZVscale-2-1}
was assessed on the test set for a range of bin numbers between 2
and 99, and compared to the results of isotonic regression and simulated
datasets (Fig.\,\ref{fig:optBin}). 

The NLL value {[}Fig.\,\ref{fig:optBin}(a){]} decreases monotonously
with increasing bin numbers up to about 40, after which, it starts
a slow noisy rise. It never crosses the isotonic regression reference
line. In contrast, the line for simulated datasets follows a monotonous
decrease. The difference between the actual NLL curve and the simulated
one results from a misfit of \emph{average calibration}, which starts
at 0 for a single bin, reaches a maximum around 40 bins and decreases
again. This feature can be observed through the $S_{cal}$ score in
{[}Fig.\,\ref{fig:optBin}(c){]}. As they share the same uncertainty
set, the \emph{sharpness} contribution is identical for both curves,
and it parallels the simulated NLL curve. Note that the minimum in
the NLL curve should not be interpreted as an optimum for the number
of bins: it points \emph{a contrario} to an area with poor calibration
performances. The areas at low and large bin numbers where the NLL
curve overlaps the simulated CI are those where the average calibration
of the scaled test set is at its best.

The $ENCE_{x}$ statistics are shown in {[}Fig.\,\ref{fig:optBin}(b){]}.
$ENCE_{u}$ decreases down to $\sim$0.1 at about 40 bins after which
it increases slightly to reach a plateau. At its minimum, it is slightly
lower than the corresponding isotonic regression reference, while
the plateau is slightly above this value. The ENCE curves for the
other binning variables present a steady decrease over the studied
bin range and they do not reach their isotonic regression reference.
In contrast to the NLL, the three $ENCE_{x}$ curves stray far above
their simulated values. 

If one considers the $S_{x}$ statistics {[}Fig.\,\ref{fig:optBin}(c){]},
the $S_{u}$ curve parallels perfectly the $ENCE_{u}$ curve. The
$S_{X_{2}}$ curve presents a shallow minimum around 70 bins, and
it gets lower than the isotonic regression value at about 40 bins.
The $S_{X_{1}}$ curve presents no minimum in the studied range. It
reaches the corresponding isotonic value near 70 bins. None of these
scores reaches the simulated values (all identical at about 0.1).
Contrasting with the other scores, $S_{cal}$ increases up to around
40 bins before decreasing. It exceeds the isotonic regression value
between 15 and 60 bins. As for the NLL curve, the $S_{cal}$ curve
matches its simulated reference at small and large bin numbers. This
behavior reflects the fact that the scaling of individual bins by
BVS cannot guarantee average calibration for the test set. The $S_{tot}$
score combines the four previous $S_{x}$ scores and shows a decrease
up to 70 bins followed by a plateau. The isotonic regression reference
is reached for about 40 bins. $S_{tot}$ is dominated by adaptivity
errors and does never match its simulated reference.

Fig.\,\ref{fig:optBin}(d) presents the $f_{v,x}$ validation statistics
with their confidence intervals. A shaded area depicts the 95\,\%
CI for the simulated datasets, identical for all three statistics.
Globally, considering the uncertainties on the statistics, there is
no noticeable departure of BVS statistics from the isotonic regression
ones, except maybe for $f_{v,u}$ between 30 and 50 bins, where it
gets closer to the 0.95 target and the simulated values. This comforts
the diagnostic that choosing $N_{B}=40$ would be the best option
to ensure consistency with NLL scaling factors.

This preliminary analysis leaves us with mixed messages: (i) it is
noticeable that most statistics are noisy and that a slight change
in bin number might affect the performance, notably when comparing
to the isotonic regression reference; (ii) if one wants to achieve
good consistency, the $ENCE_{u}$ and $S_{u}$ scores agree that one
should not use more than 20-40 bins, and in favorable cases Eq.\,\ref{eq:LZVscale-2-1}
can perform slightly better than isotonic regression; (iii) to consistently
equal or outperform isotonic regression on adaptivity scores, one
needs at least 50, and up to 80 bins. In any case, these scaling factors
do not seem to be able to achieve simultaneously consistency and adaptivity. 

However, these results do not augur of the performances of scaling
factors optimized with loss functions other than the NLL. A range
of options for alternative scaling factors will be tested in the following
for $N_{B}=20$, 40 and 80 bins. 

\subsection{BVS performance\label{subsec:BVS-performance}}

\noindent In this section, scaling factors based on loss functions
other than the NLL, i.e. $S_{tot}$, $S_{con}$ and $S_{ada}$, are
estimated and compared to the results of isotonic regression and simulated
data. The results for $N_{B}=20$, 40 and 80 are provided in Tables\,\ref{tab:Scores-and-validation-20},
\ref{tab:Scores-and-validation-40} and \ref{tab:Scores-and-validation-80},
respectively. 

I do not discuss the overall improvement brought by all scaling scenarios,
which is a trivial effect due to the improvement of average calibration,
and mostly focus on smaller scale effects related to consistency and
adaptivity. 
\begin{table}[t]
\noindent \begin{centering}
\begin{tabular}{llllr@{\extracolsep{0pt}.}llr@{\extracolsep{0pt}.}lr@{\extracolsep{0pt}.}lr@{\extracolsep{0pt}.}lr@{\extracolsep{0pt}.}lr@{\extracolsep{0pt}.}lrr@{\extracolsep{0pt}.}lr@{\extracolsep{0pt}.}lr@{\extracolsep{0pt}.}l}
\hline 
Binning & Dataset & Scaling  &  & \multicolumn{2}{c}{NLL$\downarrow$} &  & \multicolumn{2}{c}{$S_{cal}\downarrow$} & \multicolumn{2}{c}{$S_{u}\downarrow$} & \multicolumn{2}{c}{$S_{X_{1}}\downarrow$} & \multicolumn{2}{c}{$S_{X_{2}}\downarrow$} & \multicolumn{2}{c}{$S_{tot}\downarrow$} &  & \multicolumn{2}{c}{$f_{v,u_{E}}$} & \multicolumn{2}{c}{$f_{v,X_{1}}$} & \multicolumn{2}{c}{$f_{v,X_{2}}$}\tabularnewline
\hline 
 & Training & No  &  & -2&76  &  & 1&17  & 1&32  & 1&29  & 1&27  & 5&04  &  & {[}0&01, 0.09{]}  & {[}0&04, 0.17{]}  & {[}0&07, 0.20{]}\tabularnewline
\cline{2-24} 
 & Test & No  &  & -2&75  &  & 1&13  & 1&39  & 1&27  & 1&27  & 5&06  &  & {[}0&01, 0.09{]}  & {[}0&09, 0.24{]}  & {[}0&06, 0.19{]}\tabularnewline
 &  & Isotonic &  & -3&08  &  & 0&03  & 0&19  & 0&29  & 0&27  & 0&78  &  & {[}0&73, 0.89{]}  & {[}0&52, 0.71{]}  & {[}0&61, 0.79{]}\tabularnewline
 &  & Simul &  & \multicolumn{2}{c}{-} &  & \textbf{0}&\textbf{00 } & \textbf{0}&\textbf{10 } & \textbf{0}&\textbf{09 } & \textbf{0}&\textbf{09 } & \textbf{0}&\textbf{29 } &  & \textbf{{[}0}&\textbf{90,1.00{]}} & \textbf{{[}0}&\textbf{90,1.00{]}} & \textbf{{[}0}&\textbf{90,1.00{]}}\tabularnewline
\hline 
$u$ & Training & NLL  &  & \textbf{-3}&\textbf{05 } &  & \textbf{0}&\textbf{00 } & 0&21  & \textbf{0}&\textbf{33 } & \textbf{0}&\textbf{29 } & 0&83  &  & {[}0&78, 0.93{]}  & {[}0&56, 0.75{]}  & {[}0&65, 0.83{]}\tabularnewline
 &  & $S_{tot}$ &  & \textbf{-3}&\textbf{05 } &  & \textbf{0}&\textbf{00 } & 0&19  & \textbf{0}&\textbf{33 } & \textbf{0}&\textbf{29 } & 0&81  &  & {[}0&78, 0.93{]}  & {[}0&56, 0.75{]}  & {[}0&65, 0.83{]}\tabularnewline
 &  & $S_{con}$ &  & \textbf{-3}&\textbf{05 } &  & \textbf{0}&\textbf{00 } & \textbf{0}&\textbf{18 } & \textbf{0}&\textbf{33 } & \textbf{0}&\textbf{29 } & \textbf{0}&\textbf{80 } &  & {[}0&81, 0.94{]}  & {[}0&56, 0.75{]}  & {[}0&66, 0.84{]}\tabularnewline
 &  & $S_{ada}$ &  & \textbf{-3}&\textbf{05 } &  & \textbf{0}&\textbf{00 } & 0&21  & \textbf{0}&\textbf{33 } & \textbf{0}&\textbf{29 } & 0&82  &  & {[}0&80, 0.93{]}  & {[}0&56, 0.75{]}  & {[}0&65, 0.83{]}\tabularnewline
\cline{2-24} 
 & Test & NLL  &  & \textbf{-3}&\textbf{06 } &  & \textbf{0}&\textbf{03 } & 0&21  & \textbf{0}&\textbf{31 } & \textbf{0}&\textbf{27 } & 0&83  &  & {[}0&72, 0.88{]}  & {[}0&53, 0.72{]}  & {[}0&64, 0.82{]}\tabularnewline
 &  & $S_{tot}$ &  & \textbf{-3}&\textbf{06 } &  & \textbf{0}&\textbf{03 } & \textbf{0}&\textbf{20 } & \textbf{0}&\textbf{31 } & \textbf{0}&\textbf{27 } & 0&82  &  & {[}0&78, 0.93{]}  & {[}0&53, 0.72{]}  & {[}0&64, 0.82{]}\tabularnewline
 &  & $S_{con}$ &  & \textbf{-3}&\textbf{06 } &  & \textbf{0}&\textbf{03 } & \textbf{0}&\textbf{20 } & \textbf{0}&\textbf{31 } & \textbf{0}&\textbf{27 } & \textbf{0}&\textbf{81}  &  & {[}0&75, 0.90{]}  & {[}0&53, 0.72{]}  & {[}0&65, 0.83{]}\tabularnewline
 &  & $S_{ada}$ &  & \textbf{-3}&\textbf{06 } &  & \textbf{0}&\textbf{03 } & \textbf{0}&\textbf{20 } & \textbf{0}&\textbf{31 } & 0&28  & 0&82  &  & {[}0&75, 0.90{]}  & {[}0&53, 0.72{]}  & {[}0&63, 0.81{]}\tabularnewline
\hline 
$X_{1}$ & Training & NLL  &  & \textbf{-3}&\textbf{03 } &  & \textbf{0}&\textbf{00 } & 0&28  & \textbf{0}&\textbf{27 } & 0&32  & 0&87  &  & {[}0&60, 0.79{]}  & {[}0&66, 0.84{]}  & {[}0&63, 0.81{]}\tabularnewline
 &  & $S_{tot}$ &  & \textbf{-3}&\textbf{03 } &  & \textbf{0}&\textbf{00 } & \textbf{0}&\textbf{24 } & \textbf{0}&\textbf{27 } & 0&32  & \textbf{0}&\textbf{83 } &  & {[}0&67, 0.85{]}  & {[}0&65, 0.83{]}  & {[}0&62, 0.80{]}\tabularnewline
 &  & $S_{con}$ &  & \textbf{-3}&\textbf{03 } &  & \textbf{0}&\textbf{00 } & \textbf{0}&\textbf{24 } & 0&28  & 0&32  & \textbf{0}&\textbf{83 } &  & {[}0&69, 0.86{]}  & {[}0&67, 0.85{]}  & {[}0&62, 0.80{]}\tabularnewline
 &  & $S_{ada}$ &  & \textbf{-3}&\textbf{03 } &  & \textbf{0}&\textbf{00 } & 0&28  & 0&28  & \textbf{0}&\textbf{31 } & 0&87  &  & {[}0&64, 0.82{]}  & {[}0&67, 0.85{]}  & {[}0&63, 0.81{]}\tabularnewline
\cline{2-24} 
 & Test & NLL  &  & \textbf{-2}&\textbf{98 } &  & \textbf{0}&\textbf{08 } & 0&33  & 0&35  & \textbf{0}&\textbf{37 } & 1&13  &  & {[}0&52, 0.71{]}  & {[}0&51, 0.70{]}  & {[}0&52, 0.71{]}\tabularnewline
 &  & $S_{tot}$ &  & \textbf{-2}&\textbf{98 } &  & \textbf{0}&\textbf{08 } & 0&34  & 0&35  & \textbf{0}&\textbf{37 } & 1&13  &  & {[}0&49, 0.69{]}  & {[}0&53, 0.72{]}  & {[}0&51, 0.70{]}\tabularnewline
 &  & $S_{con}$ &  & \textbf{-2}&\textbf{98 } &  & \textbf{0}&\textbf{08 } & 0&33  & 0&35  & \textbf{0}&\textbf{37 } & 1&13  &  & {[}0&53, 0.72{]}  & {[}0&51, 0.70{]}  & {[}0&52, 0.71{]}\tabularnewline
 &  & $S_{ada}$ &  & \textbf{-2}&\textbf{98 } &  & \textbf{0}&\textbf{08 } & \textbf{0}&\textbf{32}  & \textbf{0}&\textbf{34}  & \textbf{0}&\textbf{37 } & \textbf{1}&\textbf{10 } &  & {[}0&53, 0.72{]}  & {[}0&51, 0.70{]}  & {[}0&52, 0.71{]}\tabularnewline
\hline 
$u\,\&\,X_{1}$ & Training & NLL &  & -3&12  &  & 0&00  & 0&11  & 0&14  & 0&17  & 0&42  &  & \textbf{{[}0}&\textbf{87, 0.98{]} } & {[}0&81, 0.94{]}  & {[}0&78, 0.93{]}\tabularnewline
 & Test &  &  & -3&03  &  & 0&17  & 0&27  & 0&24  & 0&25  & 0&93  &  & {[}0&63, 0.81{]}  & {[}0&61, 0.79{]}  & {[}0&58, 0.77{]}\tabularnewline
\hline 
O$_{x}$N$_{y}$ & Training & NLL &  & -3&03  &  & 0&00  & 0&30  & 0&31  & 0&24  & 0&85  &  & {[}0&59, 0.78{]}  & {[}0&69, 0.86{]}  & {[}0&77, 0.92{]}\tabularnewline
groups & Test &  &  & -3&01  &  & 0&04  & 0&28  & 0&32  & 0&26  & 0&91  &  & {[}0&58, 0.77{]}  & {[}0&65, 0.83{]}  & {[}0&68, 0.85{]}\tabularnewline
\hline 
\end{tabular}
\par\end{centering}
\caption{\label{tab:Scores-and-validation-20}Scores and validation metrics
for various scaling schemes with $N_{B}=20$. Bold values denote the
best scores within a group delimited by horizontal rules for the scores,
and the intervals containing the 95~\% target for the $f_{v}$ statistics.
The statistics for isotonic regression are generated from the results
of Busk \emph{et al.} \citep{Busk2022}. The ``Simul'' scores are
obtained by averaging over simulated error sets with the same uncertainties
as the training or test set.}
\end{table}
 
\begin{table}[t]
\noindent \begin{centering}
\begin{tabular}{llllr@{\extracolsep{0pt}.}llr@{\extracolsep{0pt}.}lr@{\extracolsep{0pt}.}lr@{\extracolsep{0pt}.}lr@{\extracolsep{0pt}.}lr@{\extracolsep{0pt}.}lrr@{\extracolsep{0pt}.}lr@{\extracolsep{0pt}.}lr@{\extracolsep{0pt}.}l}
\hline 
Binning & Dataset & Scaling  &  & \multicolumn{2}{c}{NLL$\downarrow$} &  & \multicolumn{2}{c}{$S_{cal}\downarrow$} & \multicolumn{2}{c}{$S_{u}\downarrow$} & \multicolumn{2}{c}{$S_{X_{1}}\downarrow$} & \multicolumn{2}{c}{$S_{X_{2}}\downarrow$} & \multicolumn{2}{c}{$S_{tot}\downarrow$} &  & \multicolumn{2}{c}{$f_{v,u_{E}}$} & \multicolumn{2}{c}{$f_{v,X_{1}}$} & \multicolumn{2}{c}{$f_{v,X_{2}}$}\tabularnewline
\hline 
 & Training & No  &  & -2&76  &  & 1&17  & 1&32  & 1&29  & 1&27  & 5&04  &  & {[}0&01, 0.09{]}  & {[}0&04, 0.17{]}  & {[}0&07, 0.20{]}\tabularnewline
\cline{2-24} 
 & Test & No  &  & -2&75  &  & 1&13  & 1&39  & 1&27  & 1&27  & 5&06  &  & {[}0&01, 0.09{]}  & {[}0&09, 0.24{]}  & {[}0&06, 0.19{]}\tabularnewline
 &  & Isotonic &  & -3&08  &  & 0&03  & 0&19  & 0&29  & 0&27  & 0&78  &  & {[}0&73, 0.89{]}  & {[}0&52, 0.71{]}  & {[}0&61, 0.79{]}\tabularnewline
 &  & Simul &  & \multicolumn{2}{c}{-} &  & \textbf{0}&\textbf{00 } & \textbf{0}&\textbf{10 } & \textbf{0}&\textbf{09 } & \textbf{0}&\textbf{09 } & \textbf{0}&\textbf{29 } &  & \textbf{{[}0}&\textbf{90,1.00{]}} & \textbf{{[}0}&\textbf{90,1.00{]}} & \textbf{{[}0}&\textbf{90,1.00{]}}\tabularnewline
\hline 
$u$ & Training & NLL  &  & \textbf{-3}&\textbf{05 } &  & \textbf{0}&\textbf{00} & 0&16  & \textbf{0}&\textbf{32 } & \textbf{0}&\textbf{28 } & 0&77  &  & {[}0&88, 0.98{]}  & {[}0&57, 0.76{]}  & {[}0&66, 0.84{]}\tabularnewline
 &  & $S_{tot}$ &  & \textbf{-3}&\textbf{05 } &  & \textbf{0}&\textbf{00} & \textbf{0}&\textbf{13 } & \textbf{0}&\textbf{32 } & \textbf{0}&\textbf{28 } & \textbf{0}&\textbf{73 } &  & {[}0&89, 0.99{]}  & {[}0&56, 0.75{]}  & {[}0&66, 0.84{]}\tabularnewline
 &  & $S_{con}$ &  & \textbf{-3}&\textbf{05 } &  & \textbf{0}&\textbf{00} & \textbf{0}&\textbf{13 } & \textbf{0}&\textbf{32 } & \textbf{0}&\textbf{28 } & \textbf{0}&\textbf{73 } &  & {[}0&87, 0.98{]}  & {[}0&57, 0.76{]}  & {[}0&66, 0.84{]}\tabularnewline
 &  & $S_{ada}$ &  & \textbf{-3}&\textbf{05 } &  & \textbf{0}&\textbf{00} & 0&17  & \textbf{0}&\textbf{32 } & \textbf{0}&\textbf{28 } & 0&77  &  & {[}0&84, 0.96{]}  & {[}0&56, 0.75{]}  & {[}0&65, 0.83{]}\tabularnewline
\cline{2-24} 
 & Test & NLL  &  & \textbf{-3}&\textbf{07 } &  & \textbf{0}&\textbf{04 } & 0&19  & \textbf{0}&\textbf{30 } & \textbf{0}&\textbf{27 } & 0&80  &  & {[}0&80, 0.93{]}  & {[}0&54, 0.73{]}  & {[}0&61, 0.79{]}\tabularnewline
 &  & $S_{tot}$ &  & \textbf{-3}&\textbf{07 } &  & \textbf{0}&\textbf{04 } & 0&19  & \textbf{0}&\textbf{30 } & \textbf{0}&\textbf{27 } & 0&80  &  & \textbf{{[}0}&\textbf{82, 0.95{]} } & {[}0&54, 0.73{]}  & {[}0&61, 0.79{]}\tabularnewline
 &  & $S_{con}$ &  & \textbf{-3}&\textbf{07 } &  & \textbf{0}&\textbf{04 } & \textbf{0}&\textbf{18 } & \textbf{0}&\textbf{30 } & \textbf{0}&\textbf{27 } & \textbf{0}&\textbf{79 } &  & \textbf{{[}0}&\textbf{83, 0.96{]} } & {[}0&54, 0.73{]}  & {[}0&61, 0.79{]}\tabularnewline
 &  & $S_{ada}$ &  & \textbf{-3}&\textbf{07 } &  & \textbf{0}&\textbf{04 } & \textbf{0}&\textbf{18 } & \textbf{0}&\textbf{30 } & \textbf{0}&\textbf{27 } & \textbf{0}&\textbf{79 } &  & \textbf{{[}0}&\textbf{82, 0.95{]} } & {[}0&55, 0.74{]}  & {[}0&61, 0.79{]}\tabularnewline
\hline 
$X_{1}$ & Training & NLL  &  & \textbf{-3}&\textbf{04 } &  & \textbf{0}&\textbf{00 } & 0&24  & \textbf{0}&\textbf{24 } & \textbf{0}&\textbf{28 } & 0&76  &  & {[}0&72, 0.88{]}  & {[}0&73, 0.89{]}  & {[}0&65, 0.83{]}\tabularnewline
 &  & $S_{tot}$ &  & \textbf{-3}&\textbf{04 } &  & \textbf{0}&\textbf{00 } & 0&21  & \textbf{0}&\textbf{24 } & \textbf{0}&\textbf{28 } & 0&73  &  & {[}0&75, 0.90{]}  & {[}0&74, 0.90{]}  & {[}0&65, 0.83{]}\tabularnewline
 &  & $S_{con}$ &  & \textbf{-3}&\textbf{04 } &  & \textbf{0}&\textbf{00 } & \textbf{0}&\textbf{20 } & \textbf{0}&\textbf{24 } & \textbf{0}&\textbf{28 } & \textbf{0}&\textbf{71 } &  & {[}0&77, 0.92{]}  & {[}0&74, 0.90{]}  & {[}0&65, 0.83{]}\tabularnewline
 &  & $S_{ada}$ &  & \textbf{-3}&\textbf{04 } &  & \textbf{0}&\textbf{00 } & 0&24  & \textbf{0}&\textbf{24 } & \textbf{0}&\textbf{28 } & 0&76  &  & {[}0&71, 0.87{]}  & {[}0&74, 0.90{]}  & {[}0&66, 0.84{]}\tabularnewline
\cline{2-24} 
 & Test & NLL  &  & \textbf{-2}&\textbf{98 } &  & \textbf{0}&\textbf{06 } & \textbf{0}&\textbf{30 } & \textbf{0}&\textbf{37 } & \textbf{0}&\textbf{38 } & \textbf{1}&\textbf{11 } &  & {[}0&60, 0.79{]}  & {[}0&48, 0.68{]}  & {[}0&49, 0.69{]}\tabularnewline
 &  & $S_{tot}$ &  & \textbf{-2}&\textbf{98 } &  & \textbf{0}&\textbf{06 } & 0&31  & \textbf{0}&\textbf{37 } & \textbf{0}&\textbf{38 } & 1&12  &  & {[}0&61, 0.79{]}  & {[}0&47, 0.67{]}  & {[}0&51, 0.70{]}\tabularnewline
 &  & $S_{con}$ &  & \textbf{-2}&\textbf{98 } &  & \textbf{0}&\textbf{06 } & \textbf{0}&\textbf{30 } & \textbf{0}&\textbf{37 } & \textbf{0}&\textbf{38 } & \textbf{1}&\textbf{11 } &  & {[}0&61, 0.79{]}  & {[}0&49, 0.69{]}  & {[}0&50, 0.70{]}\tabularnewline
 &  & $S_{ada}$ &  & \textbf{-2}&\textbf{98 } &  & \textbf{0}&\textbf{06 } & \textbf{0}&\textbf{30 } & \textbf{0}&\textbf{37 } & \textbf{0}&\textbf{38 } & 1&12  &  & {[}0&57, 0.76{]}  & {[}0&49, 0.69{]}  & {[}0&48, 0.68{]}\tabularnewline
\hline 
\end{tabular}
\par\end{centering}
\caption{\label{tab:Scores-and-validation-40}Same as Table\,\ref{tab:Scores-and-validation-20}
for $N_{B}=40$. }
\end{table}
 
\begin{table}[t]
\noindent \begin{centering}
\begin{tabular}{llllr@{\extracolsep{0pt}.}llr@{\extracolsep{0pt}.}lr@{\extracolsep{0pt}.}lr@{\extracolsep{0pt}.}lr@{\extracolsep{0pt}.}lr@{\extracolsep{0pt}.}lrr@{\extracolsep{0pt}.}lr@{\extracolsep{0pt}.}lr@{\extracolsep{0pt}.}l}
\hline 
Binning & Dataset & Scaling  &  & \multicolumn{2}{c}{NLL$\downarrow$} &  & \multicolumn{2}{c}{$S_{cal}\downarrow$} & \multicolumn{2}{c}{$S_{u}\downarrow$} & \multicolumn{2}{c}{$S_{X_{1}}\downarrow$} & \multicolumn{2}{c}{$S_{X_{2}}\downarrow$} & \multicolumn{2}{c}{$S_{tot}\downarrow$} &  & \multicolumn{2}{c}{$f_{v,u_{E}}$} & \multicolumn{2}{c}{$f_{v,X_{1}}$} & \multicolumn{2}{c}{$f_{v,X_{2}}$}\tabularnewline
\hline 
 & Training & No  &  & -2&76  &  & 1&17  & 1&32  & 1&29  & 1&27  & 5&04  &  & {[}0&01, 0.09{]}  & {[}0&04, 0.17{]}  & {[}0&07, 0.20{]}\tabularnewline
\cline{2-24} 
 & Test & No  &  & -2&75  &  & 1&13  & 1&39  & 1&27  & 1&27  & 5&06  &  & {[}0&01, 0.09{]}  & {[}0&09, 0.24{]}  & {[}0&06, 0.19{]}\tabularnewline
 &  & Isotonic &  & -3&08  &  & 0&03  & 0&19  & 0&29  & 0&27  & 0&78  &  & {[}0&73, 0.89{]}  & {[}0&52, 0.71{]}  & {[}0&61, 0.79{]}\tabularnewline
 &  & Simul &  & \multicolumn{2}{c}{-} &  & \textbf{0}&\textbf{00 } & \textbf{0}&\textbf{10 } & \textbf{0}&\textbf{09 } & \textbf{0}&\textbf{09 } & \textbf{0}&\textbf{29 } &  & \textbf{{[}0}&\textbf{90,1.00{]}} & \textbf{{[}0}&\textbf{90,1.00{]}} & \textbf{{[}0}&\textbf{90,1.00{]}}\tabularnewline
\hline 
$u$ & Training & NLL  &  & \textbf{-3}&\textbf{06 } &  & \textbf{0}&\textbf{00 } & 0&13  & 0&31  & 0&27  & 0&70  &  & \textbf{{[}0}&\textbf{89, 0.99{]} } & {[}0&56, 0.75{]}  & {[}0&66, 0.84{]}\tabularnewline
 &  & $S_{tot}$ &  & \textbf{-3}&\textbf{06 } &  & \textbf{0}&\textbf{00} & 0&10  & 0&30  & \textbf{0}&\textbf{26}  & \textbf{0}&\textbf{66 } &  & \textbf{{[}0}&\textbf{89, 0.99{]} } & {[}0&59, 0.78{]}  & {[}0&68, 0.85{]}\tabularnewline
 &  & $S_{con}$ &  & \textbf{-3}&\textbf{06 } &  & \textbf{0}&\textbf{00} & \textbf{0}&\textbf{09}  & 0&31  & 0&27  & \textbf{0}&\textbf{66 } &  & \textbf{{[}0}&\textbf{92, 1.00{]} } & {[}0&56, 0.75{]}  & {[}0&66, 0.84{]}\tabularnewline
 &  & $S_{ada}$ &  & -3&05  &  & \textbf{0}&\textbf{00} & 0&16  & \textbf{0}&\textbf{29}  & \textbf{0}&\textbf{26}  & 0&71  &  & \textbf{{[}0}&\textbf{88, 0.98{]} } & {[}0&61, 0.79{]}  & {[}0&68, 0.85{]}\tabularnewline
\cline{2-24} 
 & Test & NLL  &  & \textbf{-3}&\textbf{07 } &  & 0&02  & \textbf{0}&\textbf{22 } & 0&29  & \textbf{0}&\textbf{26 } & 0&79  &  & {[}0&63, 0.81{]}  & {[}0&56, 0.75{]}  & {[}0&64, 0.82{]}\tabularnewline
 &  & $S_{tot}$ &  & -3&06  &  & 0&01  & 0&23  & 0&29  & \textbf{0}&\textbf{26 } & \textbf{0}&\textbf{78}  &  & {[}0&71, 0.87{]}  & {[}0&57, 0.76{]}  & {[}0&63, 0.81{]}\tabularnewline
 &  & $S_{con}$ &  & \textbf{-3}&\textbf{07 } &  & 0&02  & \textbf{0}&\textbf{22 } & 0&29  & \textbf{0}&\textbf{26 } & 0&79  &  & {[}0&66, 0.84{]}  & {[}0&56, 0.75{]}  & {[}0&64, 0.82{]}\tabularnewline
 &  & $S_{ada}$ &  & -3&05  &  & \textbf{0}&\textbf{00} & 0&24  & \textbf{0}&\textbf{27 } & \textbf{0}&\textbf{26 } & \textbf{0}&\textbf{78}  &  & {[}0&64, 0.82{]}  & {[}0&61, 0.79{]}  & {[}0&61, 0.79{]}\tabularnewline
\hline 
$X_{1}$ & Training & NLL  &  & \textbf{-3}&\textbf{06 } &  & \textbf{0}&\textbf{00} & 0&20  & \textbf{0}&\textbf{14}  & 0&21  & 0&55  &  & {[}0&77, 0.92{]}  & \textbf{{[}0}&\textbf{87, 0.98{]} } & {[}0&78, 0.93{]}\tabularnewline
 &  & $S_{tot}$ &  & -3&05  &  & 0&01  & \textbf{0}&\textbf{12}  & 0&16  & \textbf{0}&\textbf{20}  & \textbf{0}&\textbf{49}  &  & \textbf{{[}0}&\textbf{87, 0.98{]} } & \textbf{{[}0}&\textbf{87, 0.98{]} } & \textbf{{[}0}&\textbf{82, 0.95{]}}\tabularnewline
 &  & $S_{con}$ &  & -3&05  &  & \textbf{0}&\textbf{00} & 0&13  & 0&15  & 0&22  & 0&50  &  & \textbf{{[}0}&\textbf{88, 0.98{]} } & \textbf{{[}0}&\textbf{86, 0.97{]} } & {[}0&78, 0.93{]}\tabularnewline
 &  & $S_{ada}$ &  & -3&05  &  & \textbf{0}&\textbf{00} & 0&19  & 0&17  & 0&22  & 0&58  &  & {[}0&81, 0.94{]}  & \textbf{{[}0}&\textbf{86, 0.97{]} } & {[}0&77, 0.92{]}\tabularnewline
\cline{2-24} 
 & Test & NLL  &  & \textbf{-2}&\textbf{97 } &  & \textbf{0}&\textbf{02 } & 0&34  & \textbf{0}&\textbf{40 } & \textbf{0}&\textbf{40 } & 1&16  &  & {[}0&53, 0.72{]}  & {[}0&52, 0.71{]}  & {[}0&50, 0.70{]}\tabularnewline
 &  & $S_{tot}$ &  & -2&96  &  & 0&03  & 0&33  & 0&42  & \textbf{0}&\textbf{40 } & 1&18  &  & {[}0&54, 0.73{]}  & {[}0&46, 0.66{]}  & {[}0&48, 0.68{]}\tabularnewline
 &  & $S_{con}$ &  & -2&96  &  & \textbf{0}&\textbf{02 } & 0&31  & \textbf{0}&\textbf{40 } & 0&41  & \textbf{1}&\textbf{15 } &  & {[}0&56, 0.75{]}  & {[}0&50, 0.70{]}  & {[}0&48, 0.68{]}\tabularnewline
 &  & $S_{ada}$ &  & -2&96  &  & \textbf{0}&\textbf{02 } & \textbf{0}&\textbf{29 } & 0&44  & 0&43  & 1&17  &  & {[}0&62, 0.80{]}  & {[}0&46, 0.66{]}  & {[}0&46, 0.66{]}\tabularnewline
\hline 
\end{tabular}
\par\end{centering}
\caption{\label{tab:Scores-and-validation-80}Same as Table\,\ref{tab:Scores-and-validation-20}
for $N_{B}=80$. }
\end{table}

\subsubsection{Uncertainty-based binning}

\noindent For all $N_{B}$ values, the $S_{tot}$ and $S_{con}$ loss
functions achieve slightly better consistency and overall scores than
NLL at the training stage. This advantage is reduced at the test stage,
but NLL never outperforms the other loss functions. No notable improvement
on adaptivity is observed for $S_{ada}$-based optimization. Globally,
the transfer to the test set leads to performances very similar to
those for the training set, except for average calibration $(S_{cal})$
which is degraded. As can be expected from the results in Fig.\,\ref{fig:optBin},
the scores are globally better at $N_{B}=40$ and 80 than at $N_{B}=20$
and the quality of the correction equals or slightly exceeds that
of isotonic regression.\textcolor{orange}{{} }Only the $f_{v,u}$ validation
statistic reaches its target at $N_{B}=40$. 

\subsubsection{$X_{1}$-based binning}

\noindent Using $X_{1}$ as a binning variable performs globally worse
than $u$-binning. Only for $N_{B}=80$ do the scores at the training
stage get notably better, with a neat improvement of the $S_{X_{1}}$
and $S_{X_{2}}$ scores and $f_{v,x}$ statistics, even over-passing
the corresponding isotonic scores for all bin numbers. There is however
some loss for the consistency scores. Besides, any improvement at
the training stage is lost when transferring to the test set, where
overall scores are consistently worse than for $u$-based binning. 

\subsubsection{2D binning}

\noindent A dual binning based on $u$ and $X_{1}$ with 20 equal-size
bins in each direction ($N_{B}=400$) was attempted for the NLL loss
function (Table\,\ref{tab:Scores-and-validation-20}). The scores
at the training stage are excellent, exceeding significantly those
of isotonic regression and getting closest to the simulated datasets.
Transfer to the test set shows contrasted performances: an improvement
over isotonic regression and $u$-based scaling for the adaptivity
scores $S_{X_{1}}$ and $S_{X_{2}}$, a small degradation for the
consistency score $S_{u}$, and more importantly a very bad average
calibration score $S_{cal}$. The overall score $S_{tot}$ is worse
than for isotonic regression.

\subsubsection{O$_{x}$N$_{y}$ group binning}

\noindent To avoid the potential problems due to stratification of
$X_{1}$ and $X_{2}$, groups were defined by their oxygen and nitrogen
content O$_{x}$N$_{y}$. To ensure a correct mapping between training
and test sets, extreme compositions had to be removed from the training
set (2 points). Also, groups in the training set containing less than
10 systems were rejected (112 points), leaving us with 23 groups with
$x\in\left\{ 0-4\right\} $ and $y\in\left\{ 0-6\right\} $. The results
are reported in Table\,\ref{tab:Scores-and-validation-20}. Globally,
the performances are slightly better than those obtained for $X_{1}$
binning, but worse than those obtained by $u$ binning, with equal
or better adaptivity metrics, but degraded consistency metrics. Except
for $S_{X_{2}}$ \textendash{} the fraction of heteroatoms $X_{2}$
being directly related to the O$_{x}$N$_{y}$ composition \textendash{}
all scores at the test stage are worse than for isotonic regression. 

\section{Discussion and conclusion\label{sec:Discussion-and-conclusion}}

\noindent The effects of bin-wise variance scaling on consistency
and adaptivity have been studied on a dataset for which isotonic regression
had previously been used, providing a reference point. The scaled
uncertainties resulting from isotonic regression present a good, but
imperfect, consistency and a problematic adaptivity. The aim of this
study was to see if BVS could improve both consistency \emph{and}
adaptivity. 

For this, I first assessed the impact of the BVS bin number $N_{B}$
on the performances of the scale factors based on the NLL loss function
and observed that it is necessary to use a large number of bins (40
to 80) to achieve scores as good or slightly better than isotonic
regression. The scores do not improve above 80 bins. This contrasts
with the use of small bin numbers ($N_{B}=15$) by Frenkel and Goldberger
\citep{Frenkel2023}.

Then, I considered several optimization and binning schemes. Optimization
was done by minimizing the NLL score or alternative $S_{x}$ loss
functions based on $z$-scores statistics. For NLL optimization, the
analytical scaling factors issued from Eq.\,\ref{eq:LZVscale-2-1}
were always optimal (as demonstrated by Frenkel and Goldberger \citep{Frenkel2023}),
while the other loss functions accepted different solutions, notably
when the number of bins was increased above 20. For the binning schemes,
I used the ``standard'' uncertainty-based binning and an alternative
binning based on molecular mass $X_{1}$. The prospect of the latter
was to improve adaptivity. A 2D binning scheme based on $u$ and $X_{1}$
and a composition-based O$_{x}$N$_{y}$ grouping scheme were also
tested for the NLL loss function.

The results show that $u$-based binning was able to reach consistency
on the training set, which could be transferred with some loss to
the test set. Nevertheless, the consistency score $S_{u}$ of the
test set is on par or slightly better than for isotonic regression
for bin numbers between 20 and 40. For larger bin numbers, this score
degrades slightly, probably due to over-parameterization. With enough
bins, the adaptivity scores can slightly improve over those of isotonic
regression. Globally however, \emph{u}-based\emph{ }BVS does not outperform
significantly isotonic regression for the QM9 dataset.

In order to reach or improve adaptivity on the training set, one needs
to use a loss function $S_{ada}$ targeting explicitly adaptivity,
and/or a $X_{1}$-based binning scheme as well as a large number of
bins. In all cases, consistency is not preserved. Besides, the scaling
factors resulting from the $X_{1}$-based binning scheme do not transfer
well to the test set, leading to the worst performances. A 2D $(u,X_{1})$
binning produced slightly better results for NLL scaling factors,
but did not improve over isotonic regression.

It was also observed that BVS scaling does not ensure average calibration
of the scaled test data (nor does isotonic regression), and that this
depends notably on the number of bins. For the present dataset, the
least degradation is observed for the smallest numbers of bins (below
6) and for the largest ones (above 70). The worst effect was however
observed for 2D binning with 400 bins, ruining the gains on adaptivity
offered by this approach.

In consequence, it seems that \emph{post hoc} calibration by BVS is
unable to reach simultaneously consistency and adaptivity. As $u$
and $X_{1}$ are not or weakly correlated, all $X_{1}$ intervals
contain similar $u$ distributions, so that a piece-wise scaling of
$u$ does not affect adaptivity more than to an average scaling effect
(no local impact). The situation is somewhat aggravated by the stratification
of the features chosen here as proxy of input features (molecular
mass and fraction of heteroatoms)\citep{Pernot2023c_arXiv}, making
the design of alternative binning schemes more complex. Group calibration
based on the N$_{x}$O$_{y}$ composition of the molecules using NLL
scaling factors did not lead to any improvement over the other schemes.\bigskip{}

\noindent The main and prospective conclusion from this study is that
it is unlikely that simple \emph{post hoc} recalibration methods such
as isotonic regression or BVS will be able to ensure both the consistency
and adaptivity levels required for reliable individual predictions.
More promising avenues are either ML-UQ methods trained on uncertainties
generated from error sets by a suitable probabilistic model, and/or
specifically designed loss functions at the learning stage\citep{Fanelli2023},
in the spirit of Approximate Bayesian Computation methods.\citep{Csillery2010,Sunnaker2013,Pernot2017}

\section*{Acknowledgments}

\noindent I warmly thank Jonas Busk for providing me all the data
for this study.

\section*{Author Declarations}

\subsection*{Conflict of Interest}

\noindent The author has no conflicts to disclose.

\section*{Code and data availability\label{sec:Code-and-data}}

\noindent The code and data to reproduce the results of this article
are available at \url{https://github.com/ppernot/2023_BVS/releases/tag/v1.2}
and at Zenodo (\url{https://doi.org/10.5281/zenodo.10018460}). The
\texttt{R},\citep{RTeam2019} \href{https://github.com/ppernot/ErrViewLib}{ErrViewLib}
package implements the validation functions used in the present study,
under version \texttt{ErrViewLib-v1.7.2} (\url{https://github.com/ppernot/ErrViewLib/releases/tag/v1.7.2}),
also available at Zenodo (\url{https://doi.org/10.5281/zenodo.8300715}).\textcolor{orange}{{} }

\bibliographystyle{unsrturlPP}
\bibliography{NN}

\begin{thebibliography}{10}

\bibitem{Tran2020}
K.~Tran, W.~Neiswanger, J.~Yoon, Q.~Zhang, E.~Xing, and Z.~W. Ulissi.
\newblock \href{http://dx.doi.org/10.1088/2632-2153/ab7e1a}{Methods for
  comparing uncertainty quantifications for material property predictions}.
\newblock {\em Mach. Learn.: Sci. Technol.}, 1:025006, 2020.

\bibitem{Guo2017}
C.~Guo, G.~Pleiss, Y.~Sun, and K.~Q. Weinberger.
\newblock \href{https://proceedings.mlr.press/v70/guo17a.html}{{On Calibration
  of Modern Neural Networks}}.
\newblock In {\em {International Conference on Machine Learning}}, pages
  1321--1330. 2017.
\newblock URL: \url{https://proceedings.mlr.press/v70/guo17a.html}.

\bibitem{Kuleshov2018}
V.~Kuleshov, N.~Fenner, and S.~Ermon.
\newblock \href{https://proceedings.mlr.press/v80/kuleshov18a.html}{Accurate
  uncertainties for deep learning using calibrated regression}.
\newblock In J.~Dy and A.~Krause, editors, {\em Proceedings of the 35th
  International Conference on Machine Learning}, volume~80 of {\em Proceedings
  of Machine Learning Research}, pages 2796--2804. PMLR, 10--15 Jul 2018.
\newblock URL: \url{https://proceedings.mlr.press/v80/kuleshov18a.html}.

\bibitem{Levi2022}
D.~Levi, L.~Gispan, N.~Giladi, and E.~Fetaya.
\newblock \href{http://dx.doi.org/10.3390/s22155540}{{Evaluating and
  Calibrating Uncertainty Prediction in Regression Tasks}}.
\newblock {\em Sensors}, 22:5540, 2022.

\bibitem{Busk2022}
J.~Busk, P.~B. J{\o}rgensen, A.~Bhowmik, M.~N. Schmidt, O.~Winther, and
  T.~Vegge.
\newblock \href{http://dx.doi.org/10.1088/2632-2153/ac3eb3}{Calibrated
  uncertainty for molecular property prediction using ensembles of message
  passing neural networks}.
\newblock {\em Mach. Learn.: Sci. Technol.}, 3:015012, 2022.

\bibitem{Angelopoulos2021}
A.~N. Angelopoulos and S.~Bates.
\newblock \href{http://dx.doi.org/10.48550/arXiv.2107.07511}{{A Gentle
  Introduction to Conformal Prediction and Distribution-Free Uncertainty
  Quantification}}.
\newblock {\em arXiv:2107.07511}, July 2021.

\bibitem{Hu2022}
Y.~Hu, J.~Musielewicz, Z.~W. Ulissi, and A.~J. Medford.
\newblock \href{http://dx.doi.org/10.1088/2632-2153/aca7b1}{{Robust and
  scalable uncertainty estimation with conformal prediction for machine-learned
  interatomic potentials}}.
\newblock {\em Mach. Learn.: Sci. Technol.}, 3:045028, November 2022.

\bibitem{Pernot2023c_arXiv}
P.~Pernot.
\newblock \href{http://dx.doi.org/10.48550/arXiv.2309.06240}{{Consistency and
  adaptivity are complementary targets for the validation of variance-based
  uncertainty quantification metrics in machine learning regression tasks}}.
\newblock {\em arXiv:2309.06240}, September 2023.

\bibitem{Reiher2022}
M.~Reiher.
\newblock
  \href{http://dx.doi.org/https://doi.org/10.1002/ijch.202100101}{Molecule-specific
  uncertainty quantification in quantum chemical studies}.
\newblock {\em Isr. J. Chem.}, 62(1-2):e202100101, 2022.

\bibitem{Pernot2023_Arxiv}
P.~Pernot.
\newblock \href{http://dx.doi.org/10.48550/arXiv.2303.07170}{{Validation of
  uncertainty quantification metrics: a primer based on the consistency and
  adaptivity concepts}}.
\newblock {\em arXiv:2303.07170}, 2023.
\newblock \href {http://arxiv.org/abs/2303.07170} {\path{arXiv:2303.07170}}.

\bibitem{Frenkel2023}
L.~Frenkel and J.~Goldberger.
\newblock \href{http://dx.doi.org/10.1109/ISBI53787.2023.10230543}{Calibration
  of a regression network based on the predictive variance with applications to
  medical images}.
\newblock In {\em {2023 IEEE 20th International Symposium on Biomedical Imaging
  (ISBI)}}, pages 1--5. IEEE, 2023.

\bibitem{Frenkel2021}
L.~Frenkel and J.~Goldberger.
\newblock \href{http://dx.doi.org/10.23919/EUSIPCO54536.2021.9616219}{{Network
  Calibration by Class-based Temperature Scaling}}.
\newblock In {\em {2021 29th European Signal Processing Conference (EUSIPCO)}},
  pages 1486--1490. IEEE, 2021.

\bibitem{Pernot2022b}
P.~Pernot.
\newblock \href{http://dx.doi.org/10.1063/5.0109572}{Prediction uncertainty
  validation for computational chemists}.
\newblock {\em J. Chem. Phys.}, 157:144103, 2022.

\bibitem{Pernot2017}
P.~Pernot and F.~Cailliez.
\newblock \href{http://dx.doi.org/10.1002/aic.15781}{A critical review of
  statistical calibration/prediction models handling data inconsistency and
  model inadequacy}.
\newblock {\em AIChE J.}, 63:4642--4665, 2017.

\bibitem{Pernot2017b}
P.~Pernot.
\newblock \href{http://dx.doi.org/10.1063/1.4994654}{The parameter uncertainty
  inflation fallacy}.
\newblock {\em J. Chem. Phys.}, 147:104102, 2017.

\bibitem{Pernot2022a}
P.~Pernot.
\newblock \href{http://dx.doi.org/10.1063/5.0084302}{The long road to
  calibrated prediction uncertainty in computational chemistry}.
\newblock {\em J. Chem. Phys.}, 156:114109, 2022.

\bibitem{Pernot2022c}
P.~Pernot.
\newblock \href{http://dx.doi.org/10.48550/arXiv.2206.15272}{{Confidence curves
  for UQ validation: probabilistic reference vs. oracle}}.
\newblock {\em arXiv:2206.15272}, June 2022.

\bibitem{Gneiting2007a}
T.~Gneiting, F.~Balabdaoui, and A.~E. Raftery.
\newblock
  \href{http://dx.doi.org/https://doi.org/10.1111/j.1467-9868.2007.00587.x}{Probabilistic
  forecasts, calibration and sharpness}.
\newblock {\em J. R. Statist. Soc. B}, 69:243--268, 2007.

\bibitem{Chung2021}
Y.~Chung, I.~Char, H.~Guo, J.~Schneider, and W.~Neiswanger.
\newblock \href{http://dx.doi.org/10.48550/arXiv.2109.10254}{Uncertainty
  toolbox: an open-source library for assessing, visualizing, and improving
  uncertainty quantification}.
\newblock {\em arXiv:2109.10254}, September 2021.

\bibitem{Pernot2023a_arXiv}
P.~Pernot.
\newblock \href{http://dx.doi.org/10.48550/arXiv.2305.11905}{Properties of the
  {ENCE} and other {MAD}-based calibration metrics}.
\newblock {\em arXiv:2305.11905}, May 2023.

\bibitem{Rasmussen2023}
M.~H. Rasmussen, C.~Duan, H.~J. Kulik, and J.~H. Jensen.
\newblock \href{http://dx.doi.org/10.26434/chemrxiv-2023-w93dm}{{Uncertain of
  uncertainties? A comparison of uncertainty quantification metrics for
  chemical data sets}}.
\newblock {\em ChemRxiv}, September 2023.

\bibitem{Watts2022}
S.~Watts and L.~Crow.
\newblock \href{http://dx.doi.org/10.48550/arXiv.2210.02848}{{The Shannon
  Entropy of a Histogram}}.
\newblock {\em arXiv:2210.02848}, October 2022.

\bibitem{DiCiccio1996}
T.~J. DiCiccio and B.~Efron.
\newblock \href{https://www.jstor.org/stable/2246110}{Bootstrap confidence
  intervals}.
\newblock {\em Statist. Sci.}, 11:189--212, 1996.
\newblock URL: \url{https://www.jstor.org/stable/2246110}.

\bibitem{Bakowies2021}
D.~Bakowies and O.~A. von Lilienfeld.
\newblock \href{http://dx.doi.org/10.1021/acs.jctc.1c00474}{Density
  {Functional} {Geometries} and {Zero}-{Point} {Energies} in {Ab} {Initio}
  {Thermochemical} {Treatments} of {Compounds} with {First}-{Row} {Atoms} ({H},
  {C}, {N}, {O}, {F})}.
\newblock {\em J. Chem. Theory Comput.}, 17:4872--4890, 2021.

\bibitem{Pernot2023b_arXiv}
P.~Pernot.
\newblock \href{http://dx.doi.org/10.48550/arXiv.2306.05180}{{Stratification of
  uncertainties recalibrated by isotonic regression and its impact on
  calibration error statistics}}.
\newblock {\em arXiv:2306.05180}, June 2023.

\bibitem{Fanelli2023}
C.~Fanelli and J.~Giroux.
\newblock \href{http://dx.doi.org/10.48550/arXiv.2310.02913}{{ELUQuant:
  Event-Level Uncertainty Quantification in Deep Inelastic Scattering}}.
\newblock {\em arXiv:2310.02913}, October 2023.

\bibitem{Csillery2010}
K.~Csill\'{e}ry, M.~G.~B. Blum, O.~E. Gaggiotti, and O.~Fran\c{c}ois.
\newblock \href{http://dx.doi.org/10.1016/j.tree.2010.04.001}{{Approximate
  Bayesian Computation (ABC) in practice}}.
\newblock {\em Trends Ecol. Evol.}, 25:410--418, 2010.

\bibitem{Sunnaker2013}
M.~Sunn{\aa}ker, A.~G. Busetto, E.~Numminen, J.~Corander, M.~Foll, and
  C.~Dessimoz.
\newblock \href{http://dx.doi.org/10.1371/journal.pcbi.1002803}{{Approximate
  Bayesian Computation}}.
\newblock {\em PLoS Comput. Biol.}, 9:e1002803, 2013.

\bibitem{RTeam2019}
{R Core Team}.
\newblock \href{http://www.R-project.org/}{{\em {R}: {A} {L}anguage and
  {E}nvironment for {S}tatistical {C}omputing}}.
\newblock R Foundation for Statistical Computing, Vienna, Austria, 2019.
\newblock URL: \url{http://www.R-project.org/}.

\end{thebibliography}

\end{document}